\documentclass{article} 
\usepackage{nips11submit_e,times}

\usepackage{algorithm}
\usepackage{algorithmic}

\usepackage{epsf}
\usepackage{amsmath}
\usepackage{amsthm}
\usepackage{graphicx}
\usepackage{color}
\usepackage{amssymb}
\usepackage{wrapfig}

\title{Regularized Off-Policy TD-Learning}

\author{
Bo Liu, \ \ Sridhar Mahadevan\\
Computer Science Department\\
University of Massachusetts\\
Amherst, MA 01003\\
\texttt{\{boliu, mahadeva\}@cs.umass.edu}
\And
Ji Liu\\
Computer Science Department\\
University of Wisconsin\\
Madison, WI 53706\\
\texttt{ji-liu@cs.wisc.edu}
}

\nipsfinalcopy 

\begin{document}

\maketitle
\begin{abstract}
We present a novel $l_1$ regularized off-policy convergent TD-learning method (termed RO-TD), which is able to learn sparse representations of value functions with low computational complexity. The algorithmic framework underlying RO-TD integrates two key ideas: off-policy convergent gradient TD methods, such as TDC, and a convex-concave saddle-point formulation of non-smooth convex optimization, which enables first-order solvers and feature selection using online convex regularization.
A detailed theoretical and experimental analysis of RO-TD is presented. A variety of experiments are presented to illustrate the off-policy convergence, sparse feature selection capability and low computational cost of the RO-TD algorithm.
\end{abstract}
\section{Introduction}
Temporal-difference (TD) learning is a widely used method in reinforcement learning (RL). Although TD converges when samples are drawn ``on-policy" by sampling from the Markov chain underlying a policy in a Markov decision process (MDP), it can be  shown to be divergent when samples are drawn ``off-policy".
Off-policy methods are of wider applications since they are able to learn while executing an exploratory policy, learn from demonstrations, and learn multiple tasks in parallel \cite{OFFACTOR:2012}.
Sutton et al. \cite{FastGradient:2009} introduced convergent off-policy temporal difference learning algorithms, such as TDC, whose computation time scales linearly with the number of samples and the number of features. Recently, a  linear off-policy actor-critic algorithm based on the same framework was proposed in \cite{OFFACTOR:2012}.

Regularizing  reinforcement learning algorithms leads to more robust methods that can scale up to large problems with many  potentially irrelevant features.
LARS-TD \cite{Kolter09LARSTD}  introduced a popular approach of combining $l_1$ regularization using  Least Angle Regression (LARS) with the least-squares TD (LSTD)   framework. Another approach was introduced in \cite{JeffLcp:nips2010}  (LCP-TD) based on the Linear Complementary Problem (LCP) formulation, an optimization approach
between linear programming and quadratic programming. LCP-TD uses ``warm-starts", which helps significantly reduce the burden of $l_1$ regularization. A theoretical analysis of $l_1$ regularization was given in \cite{LASSOTD:2011}, including
error bound analysis with finite samples in the on-policy setting.
Another approach integrating the Dantzig Selector with LSTD was proposed in \cite{DantzigRL:2012}, overcoming some of the drawbacks of LARS-TD. An approximate linear programming approach for finding $l_1$ regularized solutions of the Bellman equation was presented in \cite{ALP:2010}.  All of these approaches are second-order methods, requiring complexity approximately cubic in the number of (active) features. Another approach to feature selection is to greedily add new features, proposed recently in \cite{GreedysparseRL:2012}. Regularized first-order reinforcement learning approaches have recently been investigated in the  on-policy setting as well, wherein convergence of $l_1$ regularized temporal difference learning is discussed in \cite{parr:tech:2012} and mirror descent \cite{sra2011optimization} is used in \cite{mahadevan:MID:2012}.

In this paper, the off-policy TD learning problem is formulated from the stochastic optimization perspective. A novel objective function is proposed based on the linear equation formulation of the TDC algorithm. The  optimization problem underlying off-policy TD methods, such as TDC, is reformulated as a convex-concave saddle-point stochastic approximation problem, which is both convex and incrementally solvable. A detailed theoretical and experimental study of the RO-TD algorithm is presented.

Here is a brief roadmap to the rest of the paper.  Section~\ref{sec:Background} reviews the basics of MDPs, RL and recent work on off-policy convergent TD methods, such as  the TDC algorithm. Section~\ref{sec:MID} introduces the  proximal gradient method and the convex-concave saddle-point formulation of non-smooth convex optimization. Section~\ref{sec:Algorithm-Design} presents the new RO-TD
algorithm.  Convergence analysis of RO-TD is presented in Section~\ref{sec:ConvergenceMIrror}. Finally, in Section~\ref{sec:Empirical-Experiments},
 experimental results are presented to demonstrate the effectiveness of RO-TD.
\section{Reinforcement Learning and the TDC Algorithm\label{sec:Background}}
A  {\em Markov Decision Process} (MDP) is defined by the tuple $( {S,A,P^a_{ss'},R,\gamma } )$, comprised of a set of states $S$, a set of (possibly state-dependent) actions $A$ ($A_s$), a dynamical system model comprised of the transition kernel $P^a_{ss'}$ specifying the probability of transition to state $s'$ from state $s$ under action $a$,  a reward model $R$, and $0 \leq \gamma < 1$ is a discount factor. A policy $\pi: S \rightarrow A$ is a deterministic mapping from states to actions. Associated with each policy $\pi$ is a value function $V^\pi$, which is the fixed point of the Bellman equation:
\[{V^\pi }(s) = {T^\pi }{V^\pi }(s) = {R^\pi }(s) + \gamma {P^\pi }{V^\pi }(s)\]
where $R^\pi$ is the expected immediate reward function (treated here as a column vector) and $P^\pi$ is the state transition function under fixed policy $\pi$, and $T^\pi$ is known as the \emph{Bellman operator}. In what follows, we often drop the dependence of $V^\pi, T^\pi, R^\pi$ on $\pi$, for notational simplicity. In linear value function approximation, a value function is assumed to lie in the linear span of a basis function matrix $\Phi$ of dimension $\left| S \right| \times d$, where $d$ is the number of linear independent features. Hence, $V \approx \hat{V} = \Phi \theta$.
The vector space of all value functions is a normed inner product space, where the ``length" of any value function $f$ is measured  as $||f||_\Xi ^2 = \sum\nolimits_s \xi  (s){f^2}(s) = f'\Xi f$ weighted by $\Xi$, where $\Xi$ is defined in Figure 1.
For the $t$-th sample,  $\phi_t$,$\phi'_t$, $\theta_t$ and ${\delta _t}$ are defined in Figure 1.
TD learning uses the following update rule ${\theta _{t + 1}} = {\theta _t} + {\alpha _t}{\delta _t}{\phi _t}$, where $\alpha _t$ is the stepsize.
However, TD is only guaranteed to converge in the on-policy setting, although in many off-policy situations,  it still has satisfactory performance \cite{Kolter:offpolicyTD}.
TD with gradient correction (TDC) \cite{FastGradient:2009} aims to minimize the mean-square projected Bellman error  (MSPBE)  in order to guarantee off-policy convergence.
MSPBE is defined as
\begin{equation}\label{mspbe}
  {\rm{MSPBE}}(\theta) = \left\| {\Phi \theta - \Pi T(\Phi \theta)} \right\|_\Xi ^2
= {({\Phi ^T}\Xi (T\Phi \theta  - \Phi \theta ))^T}{({\Phi ^T}\Xi \Phi )^{ - 1}}{\Phi ^T}
  \Xi (T\Phi \theta  - \Phi \theta )
\end{equation}
To avoid computing the inverse matrix ${({\Phi ^T}\Xi \Phi )^{ - 1}}$ and to avoid the double sampling  problem \cite{sutton-barto:book}  in (\ref{mspbe}), an auxiliary variable $w$ is defined
\begin{eqnarray}
 w &=& {({\Phi ^T}\Xi \Phi )^{ - 1}}{\Phi ^T}\Xi (T\Phi \theta  - \Phi \theta )
\label{eq:w}
\end{eqnarray}
The two time-scale gradient descent learning method TDC  \cite{FastGradient:2009} is defined below
\begin{equation}
{\theta _{t + 1}} = {\theta _t} + {\alpha _t}{\delta _t}{\phi _t} - {\alpha _t}\gamma {\phi _t}^\prime (\phi _t^T{w_t}),
{w_{t + 1}} = {w_t} + {\beta _t}({\delta _t} - \phi _t^T{w_t}){\phi _t}
\label{eq:TDCupdate}
\end{equation}
where $- {\alpha _t}\gamma {\phi _t}^\prime (\phi _t^T{w_t})$ is the term for correction of gradient descent direction, and ${\beta _t} = \eta {\alpha _t},\eta  > 1$.
%
\begin{figure}[tbh]
\center
\fbox{
\begin{tabular}{p{13.4cm}}
\begin{itemize}
\item $\Xi$ is a diagonal matrix whose entries $\xi(s)$ are given by a positive probability distribution over states. $\Pi  = \Phi {({\Phi ^T}\Xi \Phi )^{ - 1}}{\Phi ^T}\Xi $ is the weighted least-squares projection operator.
\item A square root of $A$ is a matrix $B$ satisfying $B^2=A$ and $B$ is denoted as ${A^{\frac{1}{2}}}$. Note that ${A^{\frac{1}{2}}}$ may not be unique.
\item $[\cdot,\cdot]$ is a row vector, and $[\cdot;\cdot]$ is a column vector.
\item For the $t$-th sample,  $\phi_t$ (the $t$-th row of $\Phi$), $\phi'_t$ (the $t$-th row of $\Phi'$) are the feature vectors corresponding to $s_t, s'_t$, respectively.  $\theta_t$ is the coefficient vector for $t$-th sample in first-order TD learning methods, and
 ${\delta _t} = ({r_t} + \gamma \phi _t^{'T}{\theta _t}) - \phi _t^T{\theta _t}$ is the temporal difference error. Also, ${x_t} = [{w_t};{\theta _t}]$, $\alpha_t$ is a stepsize, ${\beta _t} = \eta {\alpha _t}, \eta>0$.
\item $m,n$ are conjugate numbers if $\frac{1}{m} + \frac{1}{n} = 1,m \ge 1,n \ge 1$.
$||x||{_m} = {(\sum\nolimits_j {|{x_j}{|^m}} )^{\frac{1}{m}}}$ is the $m$-norm of vector $x$.
\item $\rho$ is $l_1$ regularization parameter, $\lambda$ is the eligibility trace factor, $N$ is the sample size, $d$ is the number of basis functions, $p$ is the number of active basis functions.
\end{itemize}
\end{tabular}}
\caption{Notation used in this paper.}
\label{fig:notationmsma}
\end{figure}
\vspace{-0.5cm}
\section{Proximal Gradient  and Saddle-Point First-Order Algorithms\label{sec:MID}}
We now introduce some background material from convex optimization.
%
%
The proximal mapping associated with a convex function $h$ is defined as:\footnote{The proximal mapping can be shown to be the resolvent of the subdifferential of the function $h$.}
\begin{equation}
prox_{h}(x)=\arg\mathop{\min}\limits _{u}(h(u)+\frac{1}{2}{\left\Vert {u-x}\right\Vert ^{2}})
\label{eq:0}
\end{equation}
In the case of $h(x) = \rho {\left\| x \right\|_1} (\rho>0)$, which is particularly important for sparse feature selection, the proximal operator turns out to be the soft-thresholding
operator ${S_\rho }( \cdot )$, which is an \emph{entry-wise} shrinkage operator:
\begin{equation}
pro{x_h}{(x)_i} = {S_\rho }({x_i}) = \max ({x_i} - \rho ,0) - \max ( - {x_i} - \rho ,0)
\end{equation}
where $i$ is the index, and $\rho$ is a threshold. With this background, we now introduce the proximal gradient method.
If the optimization problem is
\begin{equation}
{x^*} = \arg \mathop {\min }\limits_{x \in X} \left( {f(x) + h(x)} \right)
\label{eq:1gw}
\end{equation}
wherein  $f(x)$ is a convex and differentiable loss function and  the regularization term $h(x)$ is convex, possibly non-differentiable and computing $prox_h$ is not expensive, then computation of (\ref{eq:1gw}) can be carried out using the \emph{proximal gradient method}:
\begin{equation}\label{linearproximal}
{x_{t + 1}} = pro{x_{{\alpha _t}h}}\left( {{x_t} - {\alpha _t}\nabla f({x_t})} \right)
\end{equation}
where ${\alpha _t} > 0$ is a (decaying) stepsize, a constant or it can be determined by line search.
\subsection{Convex-concave Saddle-Point First Order Algorithms \label{sec:saddle}}
The key  novel contribution of our paper is a convex-concave saddle-point formulation for regularized off-policy TD learning.
A convex-concave saddle-point problem is formulated as follows. Let
$x\in X,y\in Y$,  where $X,Y$ are both nonempty bounded closed convex sets,  and $f(x):X\to\mathbb{R}$
be a convex function. If there exists a function $\varphi(\cdot,\cdot)$
such that $f(x)$ can be represented as $f(x): = {\sup _{y \in Y}}\varphi (x,y)$,
then the pair $(\varphi,Y)$ is referred as the saddle-point representation
of $f$. The optimization problem
of minimizing $f$ over $X$ is converted into an equivalent convex-concave
saddle-point problem
$SadVal = {\inf _{x \in X}}{\sup _{y \in Y}}\varphi (x,y)$
of $\varphi$ on $X\times Y$. If $f$ is non-smooth
yet convex and well structured, which is not suitable for many existing
optimization approaches requiring smoothness, its saddle-point representation
$\varphi$ is often smooth and convex. Thus, convex-concave saddle-point
problems are, therefore, usually better suited for first-order methods \cite{sra2011optimization}. A comprehensive overview on extending convex minimization to convex-concave saddle-point problems with unified variational inequalities is presented in \cite{ben2005non}.
As an example, consider $f(x) = ||Ax - b|{|_m}$ which admits a bilinear minimax representation
\begin{equation}
f(x): = {\left\| {Ax - b} \right\|_m} = {\max _{{{\left\| y \right\|}_n} \le 1}}{y^T}(Ax - b)
\label{eq:minimax}
\end{equation}
where $m,n$ are conjugate numbers. Using the approach in \cite{RobustSA:2009}, Equation (\ref{eq:minimax}) can be solved as
\begin{equation}
{x_{t + 1}} = {x_t} - {\alpha _t}{A^T}{y_t},{y_{t + 1}} = {\Pi _n}({y_t} + {\alpha _t}(A{x_t} - b))
\label{eq:minimaxFOM}
\end{equation}
where $\Pi_n$ is the projection operator of $y$ onto the unit $l_n$-ball  ${\left\| y \right\|_n} \le {\rm{1}}$,which is defined as
\begin{equation}\label{l2proj}
{\Pi _n}(y) = \min (1,1/{\left\| y \right\|_n})y,n = 2,3, \cdots ,{\Pi _\infty }{\rm{ }}({y_i}) = \min (1,1/|{y_i}|){y_i}
\end{equation}
and $\Pi _\infty$ is an entrywise operator.
\section{Regularized Off-policy Convergent TD-Learning\label{sec:Algorithm-Design}}

We now describe a novel algorithm, regularized off-policy convergent TD-learning (RO-TD), which combines off-policy convergence and scalability to large feature spaces.
The objective function is proposed based on the linear equation formulation of the TDC algorithm. Then the objective function is represented via its dual minimax problem.
The RO-TD algorithm is proposed based on the primal-dual subgradient saddle-point algorithm, and inspired by related methods in  \cite{nedic2009subgradient},\cite{RobustSA:2009}.

%
\subsection{Objective Function of Off-policy TD Learning}
In this subsection, we describe the objective function of the regularized off-policy RL problem.
We now first formulate the two updates of $\theta_t, w_t$ into a single iteration by rearranging the two equations in (\ref{eq:TDCupdate}) as ${x_{t + 1}} = {x_t} - {\alpha _t}({A_t}{x_t}  - {b_t})$, where $x_t = [w_t; \theta_t]$,
\begin{equation}
{A_t} = \left[ {\begin{array}{*{20}{c}}
{\eta {\phi _t}{\phi _t}^T}&{\eta {\phi _t}{{({\phi _t} - \gamma {{\phi '}_t})}^T}}\\
{\gamma {{\phi '}_t}{\phi _t}^T}&{{\phi _t}{{({\phi _t} - \gamma {{\phi '}_t})}^T}}
\end{array}} \right],{b_t} = {\rm{ }}\left[ {\begin{array}{*{20}{c}}
{\eta {r_t}{\phi _t}}\\
{{r_t}{\phi _t}}
\end{array}} \right]
\label{eq:abc}
\end{equation}
Following \cite{FastGradient:2009},  the TDC algorithm solution follows from the linear equation $Ax = b$, where
\begin{equation}\label{eq:UNIFY}
A = \mathbb{E}[{A_t}],  b = \mathbb{E}[{b_t}],  x=[w; \theta]
\end{equation}
There are some issues regarding the objective function, which arise from the online convex optimization and reinforcement learning perspectives, respectively.
The first concern is that the objective function should be convex and stochastically solvable. Note that $A, A_t$ are neither PSD nor symmetric, and it is not straightforward to formulate a convex objective function based on them. The second concern is that since we do not have knowledge of $A$, the objective function should be separable so that it is stochastically solvable based on $A_t, b_t$.
The other concern regards the sampling condition in temporal difference learning: double-sampling. As pointed out in \cite{sutton-barto:book}, double-sampling is a necessary condition to obtain an unbiased estimator if the objective function is the Bellman residual or its derivatives (such as projected Bellman residual), wherein the product of Bellman error or projected Bellman error metrics are involved. To overcome this sampling condition constraint, the product of TD errors should be avoided in the computation of gradients. Consequently, based on the linear equation formulation in (\ref{eq:UNIFY}) and the requirement on the objective function discussed above, we propose the regularized loss function as
\begin{equation}\label{eq:lossfunc}
L(x) = {\left\| {Ax - b} \right\|_m} + h(x)
\end{equation}

Here we also enumerate some intuitive objective functions and give a brief analysis on the reasons why they are not suitable for regularized off-policy first-order TD learning.
One intuitive idea is to add a sparsity penalty on MSPBE, i.e.,
$L(\theta) = {\rm{MSPBE(}}\theta {\rm{) + }}\rho {\left\| \theta  \right\|_1}.$
Because of the $l_1$ penalty term, the solution to $\nabla L = 0$ does not have an analytical form and is thus difficult to compute.
The second intuition is to use the online least squares formulation of the linear equation
$Ax=b$. However, since $A$ is not symmetric and positive semi-definite (PSD), ${A^{\frac{1}{2}}}$ does not exist and thus $Ax=b$ cannot be reformulated as ${\min _{x \in X}}|| {{A^{\frac{1}{2}}}x - {A^{ - \frac{1}{2}}}b} ||_2^2$.
Another possible idea is to attempt to find an objective function whose gradient is exactly  ${A_t}{x_t}-{b_t}$ and thus the regularized gradient is $pro{x_{{\alpha _t}h({x_t})}}({A_t}{x_t}-{b_t})$. However, since $A_t$ is not symmetric, this gradient does not explicitly correspond to any kind of optimization problem, not to mention a convex one{\footnote{Note that the $A$ matrix in GTD2's linear equation representation is symmetric, yet is not PSD, so  it cannot be formulated as a convex problem.}}.
\subsection{RO-TD Algorithm Design \label{sec:stochasticsaddlepoint}}
In this section, the problem of (\ref{eq:lossfunc}) is formulated as a convex-concave saddle-point problem, and the RO-TD algorithm is proposed. Analogous to (\ref{eq:minimax}), the regularized
loss function  can be formulated as
\begin{equation}
{\left\| {Ax - b} \right\|_m} + h(x) = {\max _{{{\left\| y \right\|}_n} \le 1}}{y^T}(Ax - b) + h(x)
\label{eq:regminimax}
\end{equation}
Similar to (\ref{eq:minimaxFOM}), Equation (\ref{eq:regminimax}) can be solved via an iteration procedure as follows, where ${{ x}_t} = \left[ {{w _t};{\theta_t}} \right]$.
\begin{eqnarray}
\nonumber
{{ x}_{t + \frac{1}{2}}}
= {x_t} - {\alpha _t}{A_{t}^T}{y_t}
&,&
{{ y}_{t + \frac{1}{2}}}
 = { y_t} + \alpha_t ({A_t}{x_t} - {b_t})\\
{x_{t + 1}} = pro{x_{{\alpha _t}h}}({x_{t + \frac{1}{2}}})
&,&
{y_{t + 1}} = {\Pi _n}({y_{t + \frac{1}{2}}})
\label{eq:RO-TD}
\end{eqnarray}
The averaging step, which plays a crucial role in stochastic optimization convergence, generates the \emph{approximate saddle-points} \cite{sra2011optimization, nedic2009subgradient}
\begin{equation}\label{eq:averaging}
{\bar x_{t}} = {\left( {\sum\nolimits_{i = 0}^t {{\alpha _i}} } \right)^{ - 1}}\sum\nolimits_{i = 0}^t {{\alpha _i}{x_i}} ,{\bar y_{t}} = {\left( {\sum\nolimits_{i = 0}^t {{\alpha _i}} } \right)^{ - 1}}\sum\nolimits_{i = 0}^t {{\alpha _i}{y_i}}
\end{equation}
Due to the computation of $A_t$ in (\ref{eq:RO-TD}) at each iteration, the computation cost appears to be $O(Nd^2)$, where $N, d$ are defined in Figure \ref{fig:notationmsma}.
However, the computation cost is actually $O(Nd)$ with a linear algebraic trick by computing not $A_t$ but $y_t^T{A_t} ,{A_t}{x_t} - {b_t}$. Denoting ${y_t} = [{y_{1,t}};{y_{2,t}}]$, where ${y_{1,t}};{y_{2,t}}$ are column vectors of equal length, we have
\begin{equation}\label{eq:ytAt}
y_t^T{A_t} = \left[ {\begin{array}{*{20}{c}}
{{\eta \phi _t^T(y_{1,t}^T{\phi _t}) + \gamma \phi _t^T(y_{2,t}^T \phi _t^\prime )}}
&{{{{({\phi _t} - \gamma \phi _t^\prime )}^T}(\eta y_{1,t}^T + y_{2,t}^T){\phi _t}}}
\end{array}} \right]
\end{equation}
${A_t}{x_t} - {b_t}$ can be computed according to Equation (\ref{eq:TDCupdate}) as follows:
\begin{equation}\label{eq:Atxt-bt}
{A_t}{x_t} - {b_t} = \left[ {\begin{array}{*{20}{c}}
{ - \eta ({\delta _t} - \phi _t^T{w_t}){\phi _t}};
{\gamma  (\phi _t^T{w_t}){\phi _t}^\prime - {\delta _t}{\phi _t}}
\end{array}} \right]
\end{equation}
Both  (\ref{eq:ytAt}) and (\ref{eq:Atxt-bt}) are of linear computation complexity. Now we are ready to present the RO-TD algorithm:
\begin{algorithm}
\label{alg: RO-TD}
\caption{RO-TD}
Let $\pi$ be some fixed policy of an MDP $M$, and let the sample set $S = \{{s_{i}},{r_{i}},{s_{i}}'\}_{i=1}^{N}$.
Let $\Phi$ be some fixed basis.
\begin{algorithmic}[1]
\REPEAT
\STATE Compute ${{\phi_{t}},{\phi_{t}}'}$ and TD error
${\delta_{t}} = ({r_{t}}+\gamma\phi_{t}^{'T}{\theta_{t}})-\phi_{t}^{T}{\theta_{t}}$
\STATE Compute $y_{_t}^T{A_t},{A_t}{x_t} - {b_t}$ in Equation (\ref{eq:ytAt}) and (\ref{eq:Atxt-bt}).
\STATE Compute $x_{t+1}, y_{t+1}$ as in Equation (\ref{eq:RO-TD})
\STATE Set $t\leftarrow t+1$;
\UNTIL {$t=N$};
\STATE Compute $\bar x_{N}, \bar y_{N}$ as in Equation (\ref{eq:averaging}) with $t=N$
\end{algorithmic}
\end{algorithm}

There are some design details of the algorithm to be elaborated. First, the regularization term $h(x)$ can be any kind of convex regularization, such as ridge regression or sparsity penalty $\rho||x||_1$. In case of $h(x) = \rho ||x||_1$, $pro{x_{{\alpha _t}h}}( \cdot ) = {S_{{\alpha _t}\rho }}( \cdot )$.
In real applications the sparsification requirement on $\theta$ and auxiliary variable $w$ may be different, i.e., $h(x) = {\rho _1}{\left\| \theta  \right\|_1} + {\rho _2}{\left\| w \right\|_1},{\rho _1} \ne {\rho _2}$, one can simply replace the uniform soft thresholding $S_{{\alpha _t}\rho}$ by two separate soft thresholding operations $S_{{\alpha _t}\rho_1}, S_{{\alpha _t}\rho_2}$ and thus the third equation in (\ref{eq:RO-TD}) is replaced by the following,
\begin{equation}\label{twothreshold}
{x_{t + \frac{1}{2}}} = \left[ {{w_{t + \frac{1}{2}}};{\theta_{t + \frac{1}{2}}}} \right], {\theta _{t + 1}} = {S_{{\alpha _t}{\rho _1}}}({\theta _{t + \frac{1}{2}}}),{w_{t + 1}} = {S_{{\alpha _t}{\rho _2}}}({w_{t + \frac{1}{2}}})
\end{equation}
Another concern is the choice of conjugate numbers $(m,n)$. For ease of computing $\Pi_n$, we use $(2,2)$($l_2$ fit), $(+\infty, 1)$(uniform fit) or $(1,+\infty)$. $m=n=2$ is used in the experiments below.
\subsection{ RO-GQ($\lambda$) Design}
GQ($\lambda$)\cite{gq:maei2010} is a  generalization of the TDC algorithm with eligibility traces and off-policy learning of temporally abstract predictions, where the gradient update changes from Equation (\ref{eq:TDCupdate}) to
\begin{equation}\label{eq:gq}
{\theta _{t + 1}} = {\theta _t} + {\alpha _t}[{\delta _t}{e_t} - \gamma (1 - \lambda ){w_t}^T{e_t}{\bar \phi _{t + 1}}],{w_{t + 1}} = {w_t} + {\beta _t}({\delta _t}{e_t} - w_t^T{\phi _t}{\phi _t})
\end{equation}
The central element is to extend the MSPBE function to the case where it incorporates eligibility traces. The objective function and corresponding linear equation component $A_t,b_t$ can be written as follows:
\begin{equation}\label{eq:gqObj}
L(\theta ) = ||\Phi \theta  - \Pi T ^{\pi \lambda}\Phi \theta ||_\Xi ^2
\end{equation}
%
\begin{equation}\label{eq:gqAb}
{A_t} = \left[ {\begin{array}{*{20}{c}}
{\eta {\phi _t}{\phi _t}^T}&{\eta {e_t}{{({\phi _t} - \gamma {\bar \phi _{t + 1}})}^T}}\\
{\gamma (1 - \lambda ){{\bar \phi }_{t + 1}}e_t^T}&{{e_t}{{({\phi _t} - \gamma {\bar \phi _{t + 1}})}^T}}
\end{array}} \right],{b_t} = \left[ {\begin{array}{*{20}{c}}
{\eta {r_t}{e_t}}\\
{{r_t}{e_t}}
\end{array}} \right]
\end{equation}
Similar to Equation (\ref{eq:ytAt}) and (\ref{eq:Atxt-bt}), the computation of $y_{_t}^T{A_t},{A_t}{x_t} - {b_t}$ is
\begin{eqnarray}\label{eq:gqcase}
\nonumber
y_{_t}^T{A_t}
&=&\left[ {\begin{array}{*{20}{c}}
{\eta \phi _t^T(y_{1,t}^T{\phi _t}) + \gamma (1 - \lambda )e_t^T(y_{2,t}^T{\bar \phi _{t + 1}})}
&{{({\phi _t} - \gamma {\bar \phi _{t + 1}})^T}(\eta y_{1,t}^T + y_{2,t}^T){e_t}}
\end{array}} \right]\\
{A_t}{x_t} - {b_t} &=& \left[ {\begin{array}{*{20}{c}}
{ - \eta ({\delta _t}{e_t} - \phi _t^T{w_t}{\phi _t})};
{\gamma (1 - \lambda )(e_t^T{w_t}){\bar \phi _{t + 1}} - {\delta _t}{e_t}}
\end{array}} \right]
\end{eqnarray}
where eligibility traces $e_t$, and ${\bar \phi _{t }},  T ^{\pi \lambda}$  are defined in \cite{gq:maei2010}.
Algorithm 2, RO-GQ($\lambda$), extends the RO-TD algorithm to include eligibility traces.
\vspace{-0.3cm}
\begin{algorithm}
\label{alg: RO-GQ}
\caption{RO-GQ($\lambda$)}
Let $\pi$ and $\Phi$ be as defined in Algorithm 1. Starting from $s_0$.
\begin{algorithmic}[1]
\REPEAT
\STATE Compute ${{\phi_{t}},{\bar \phi_{t+1}}}$ and TD error
${\delta_{t}} = ({r_{t}}+\gamma \bar \phi_{t+1}^{T}{\theta_{t}})-\phi_{t}^{T}{\theta_{t}}$
\STATE  Compute $y_{_t}^T{A_t},{A_t}{x_t} - {b_t}$ in Equation (\ref{eq:gqcase}).
\STATE  Compute $x_{t+1}, y_{t+1}$ as in Equation (\ref{eq:RO-TD})
\STATE  Choose action $a_t$, and get $s_{t+1}$
\STATE  Set $t\leftarrow t+1$;
\UNTIL {$s_t$ is an absorbing state};
\STATE Compute $\bar x_{t}, \bar y_{t}$ as in Equation (\ref{eq:averaging})
\end{algorithmic}
\end{algorithm}
\subsection{Extension \label{sec:Extension}}
It is also worth noting that there exists another formulation of the loss
function different from Equation (\ref{eq:lossfunc}) with the following convex-concave
formulation as  in \cite{nesterov:composite:2007, sra2011optimization},
\begin{eqnarray}
\nonumber
\mathop {\min }\limits_x \frac{1}{2}\left\| {Ax - b} \right\|_2^2 + \rho {\left\| x \right\|_1}
 &=& \mathop {\max }\limits_{{{\left\| {{A^T}y} \right\|}_\infty } \le 1} ({b^T}y - \frac{\rho }{2}{y^T}y) \\
 &=& \mathop {\min }\limits_x \mathop {\max }\limits_{{{\left\| u \right\|}_\infty } \le 1,y} \left( {{x^T}u + {y^T}(Ax - b) - \frac{\rho }{2}{y^T}y} \right)
\label{eq:formula2}
\end{eqnarray}
%
which can be solved iteratively without the proximal gradient step as follows, which serves as a counterpart of Equation (\ref{eq:RO-TD}),
\begin{eqnarray}
\nonumber
{x_{t + 1}} = {x_t} - {\alpha _t}\rho ({u_t} + {A_t}^T{y_t})      &,&     {y_{t + 1}} = {y_t} + \frac{{{\alpha _t}}}{\rho }({A_t}{x_t} - {b_t} - \rho {y_t})        \\
{u_{t + \frac{1}{2}}} = {u_t} + \frac{{{\alpha _t}}}{\rho }{x_t}  &,&      {u_{t + 1}} = {\Pi _\infty }({u_{t + \frac{1}{2}}})
\end{eqnarray}
\label{eq:ROTD2}

\section{Convergence Analysis of RO-TD \label{sec:ConvergenceMIrror}}
\textbf{Assumption 1} (\textbf{MDP})\cite{FastGradient:2009}:
The underlying Markov Reward Process (MRP) $M=(S,P,R,\gamma)$ is finite and mixing, with stationary distribution $\pi$.
Assume that $\exists \mbox{ a scalar } {R_{\max}} \mbox{ such that } {\rm {}}Var[{r_{t}}|{s_{t}}]\le{R_{\max}}$ holds w.p.1.
\\
\textbf{Assumption 2} (\textbf{Basis Function})\cite{FastGradient:2009}: $\Phi$ is a full column rank matrix, namely, $\Phi$ comprises a linear independent set of basis functions w.r.t all sample states in sample set $S$. Also, assume the features $(\phi_t,\phi_{t}^{'})$ have uniformly bounded second moments.
Finally, if $({s_{t}},{a_{t}},s_{t}^{'})$ is an i.i.d sequence, $\forall t,{\left\| {{\phi _t}} \right\|_\infty } <  + \infty ,{\left\| {\phi {'_t}} \right\|_\infty } <  + \infty$.
\\
\textbf{Assumption 3} (\textbf{Subgradient Boundedness})\cite{nedic2009subgradient}: Assume for the bilinear convex-concave loss function  defined in (\ref{eq:regminimax}),   the sets $X,Y$ are closed compact sets. Then the subgradient $y_{_t}^T{A_t}$ and ${A_t}{x_t} - {b_t}$ in RO-TD algorithm are uniformly bounded, i.e., there exists a constant $L$  such that $\left\| {{A_t}{x_t} - {b_t}} \right\| \le L,\left\| {y_{_t}^T{A_t}} \right\| \le L$.

\noindent\textbf{ Proposition 1:} The approximate saddle-point $\bar x_t$ of RO-TD converges w.p.1 to the global minimizer of the following,
\begin{equation}\label{TH1}
{x^*} = \arg {\min _{x \in X}}{\left\| {Ax - b} \right\|_m} + \rho {\left\| x \right\|_1}
\end{equation}
\\
\textbf{Proof Sketch}: See the supplementary material for details.
\section{Empirical Results\label{sec:Empirical-Experiments}}
We now demonstrate the effectiveness of the RO-TD algorithm against other algorithms across a number of benchmark domains. LARS-TD \cite{Kolter09LARSTD}, which is a popular second-order sparse reinforcement learning algorithm, is used as the baseline algorithm for feature selection and TDC is used as the off-policy convergent RL baseline algorithm, respectively.
\subsection{MSPBE Minimization and Off-Policy Convergence}
\begin{figure}
\centering
\begin{minipage}{1\textwidth}
\includegraphics[width= .32\textwidth, height=1.5in]{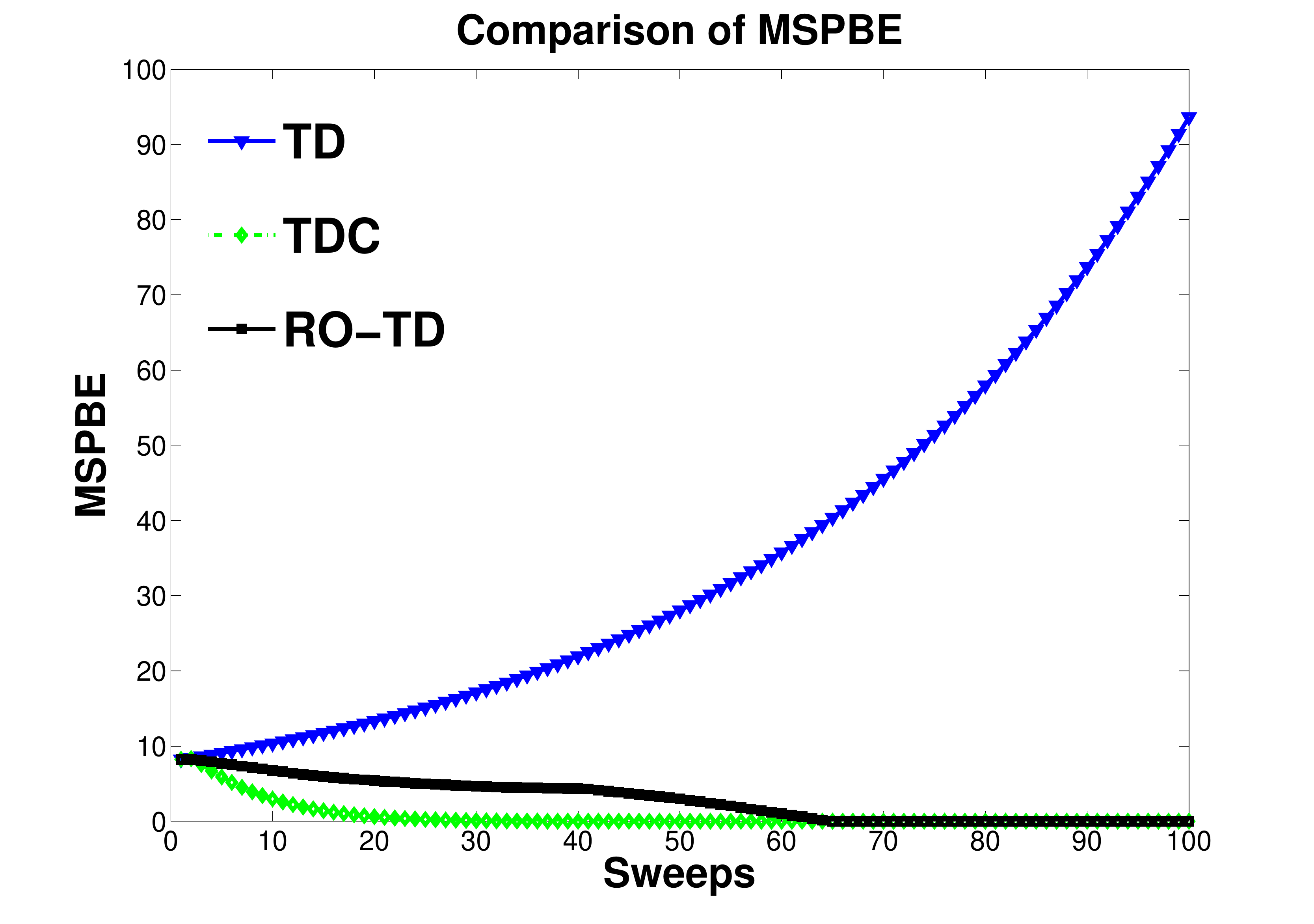}
\includegraphics[width= .32\textwidth, height=1.5in]{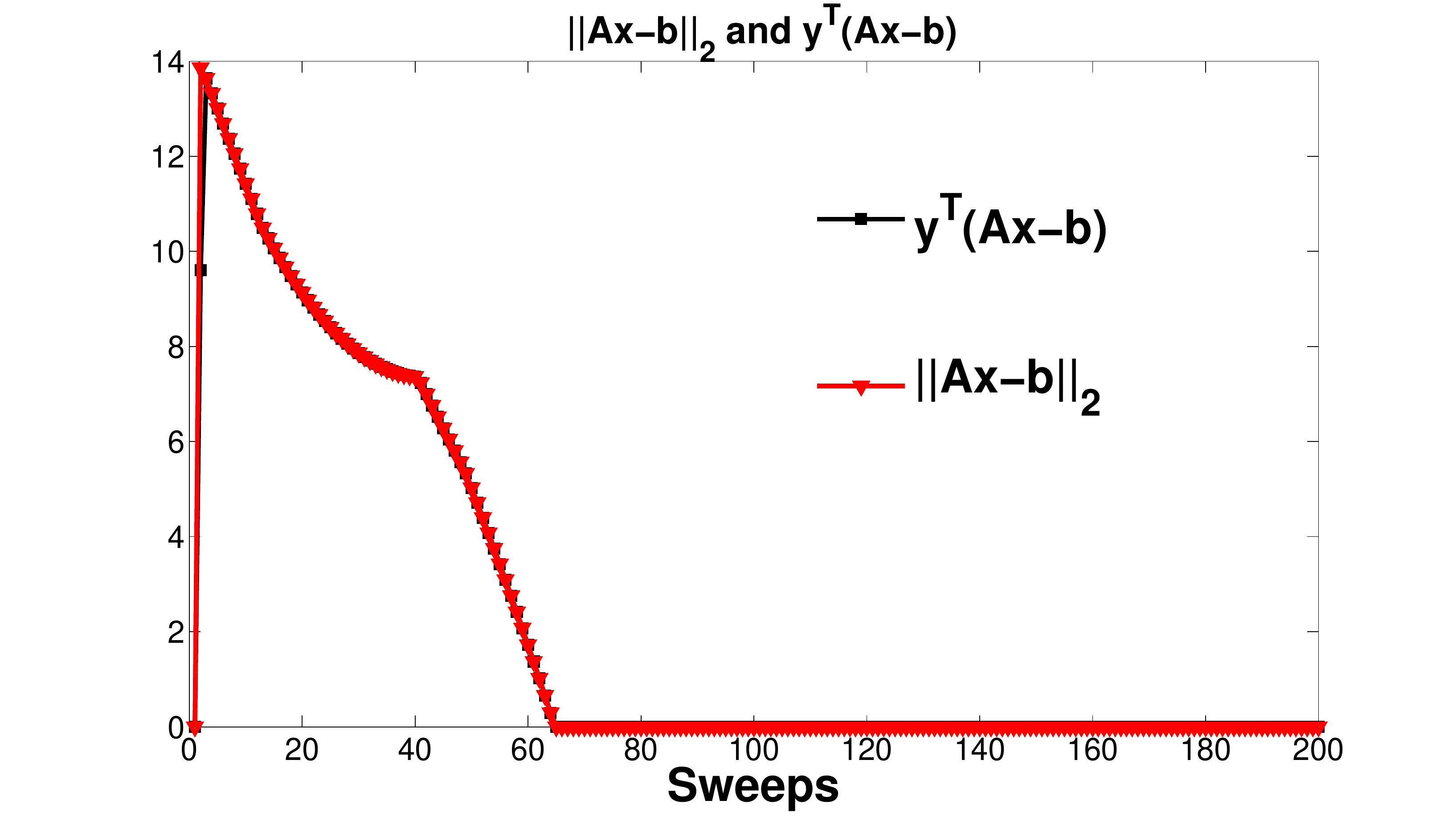}
\includegraphics[width= .32\textwidth, height=1.5in]{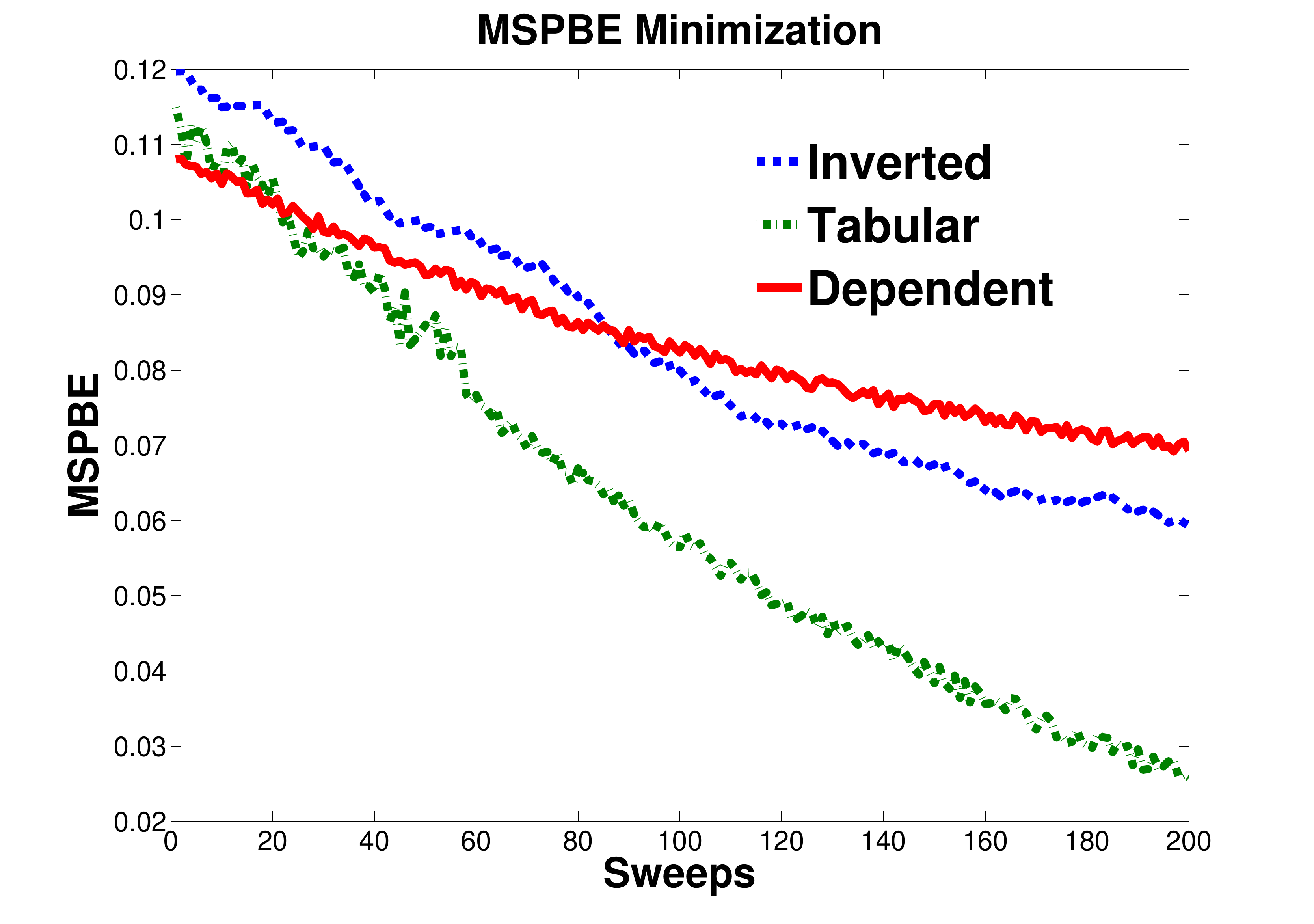}
\caption{Illustrative examples of the convergence of RO-TD using the Star and Random-walk MDPs.}\label{fig:STAR}
\end{minipage}
\end{figure}
This experiment aims to show the minimization of  MSPBE and off-policy convergence of the RO-TD algorithm. The $7$ state star MDP is a well known counterexample where TD diverges monotonically and TDC converges. It consists of $7$ states and the reward w.r.t any transition is zero. Because of this,  the star MDP is unsuitable for LSTD-based algorithms, including LARS-TD since ${\Phi ^T}R = 0$ always holds. The random-walk problem  is a standard Markov chain with $5$ states and two absorbing state at two ends. Three sets of different bases $\Phi$ are used in \cite{FastGradient:2009}, which are tabular features, inverted features and dependent features respectively.
An identical experiment setting to \cite{FastGradient:2009} is used for these two domains. The regularization term $h(x)$ is set to $0$ to make a fair comparison with TD and TDC. $\alpha = 0.01$, $\eta = 10$ for TD, TDC and RO-TD.
The comparison with TD, TDC and RO-TD is shown in the left subfigure of Figure \ref{fig:STAR}, where TDC and RO-TD have almost identical MSPBE over iterations. The middle subfigure shows the value of $y_{_t}^T(A{x_t} - b)$ and ${\left\| {A{x_t} - b} \right\|_2}$, wherein ${\left\| {A{x_t} - b} \right\|_2}$ is always greater than the value of $y_{_t}^T(A{x_t} - b)$. Note that for this problem, the Slater condition is satisfied so there is no duality gap between the two curves. As the result shows, TDC and RO-TD perform equally well, which illustrates the off-policy convergence of the RO-TD algorithm.
The result of random-walk chain is averaged over $50$ runs. The rightmost subfigure of Figure \ref{fig:STAR} shows that RO-TD is able to reduce MSPBE over successive iterations w.r.t three different basis functions.
\subsection{Feature Selection}
In this section, we use the  mountain car example with a variety of bases to show the feature selection capability of RO-TD.
The Mountain car MDPis an optimal control problem with a continuous two-dimensional state space. The steep discontinuity in the value function makes learning difficult for bases with global support. To make a fair comparison, we use the same basis function setting as in \cite{Kolter09LARSTD}, where two dimensional grids of $2,4,8,16,32$ RBFs are used so that there are totally $1365$ basis functions. For LARS-TD, $500$ samples are used. For RO-TD and TDC, $3000$ samples are used by executing $15$ episodes with $200$ steps for each episode, stepsize $\alpha_t=0.001$, and $\rho_1=0.01,\rho_2=0.2$. We use the result of LARS-TD and $l_2$ LSTD reported in \cite{Kolter09LARSTD}.
As the result shows in Table~\ref{tab:mcar}, RO-TD is able to perform feature selection successfully, whereas TDC and TD failed. It is worth noting that comparing the performance of RO-TD and LARS-TD is not the focus of this paper since LARS-TD is not convergent off-policy and RO-TD's performance can be further optimized using the mirror-descent approach  with the Mirror-Prox algorithm \cite{sra2011optimization} which incorporates mirror descent with an extragradient \cite{extragradient:1976}, as discussed below.

%
\begin{table}[h]
\centering
\begin{tabular}{|c|c|c|c|c|c|}
\hline
Algorithm & LARS-TD     & RO-TD  & $l_{2}$ LSTD & TDC & TD\tabularnewline
\hline
Success($20/20$) & $100\%$ & $100\%$ & $0\%$ & $0\%$ & $0\%$ \tabularnewline
\hline
Steps & $142.25\pm9.74$ & $147.40\pm13.31$ & - & - & - \tabularnewline
\hline
\end{tabular}
\caption{Comparison of TD, LARS-TD, RO-TD, $l_2$ LSTD, TDC and TD}\label{tab:mcar}
\end{table}
\begin{table}
\centering
\begin{tabular}{|c|c|c| c|}
\hline
Experiment\textbackslash Method & RO-GQ($\lambda$) & GQ($\lambda$)   & LARS-TD \tabularnewline
\hline
Experiment 1                      & $6.9 \pm 4.82$  &  $11.3 \pm 9.58$   & - \tabularnewline
\hline
Experiment 2                      & $14.7 \pm 10.70$ &  $27.2 \pm 6.52$  & - \tabularnewline
\hline
\end{tabular}
\caption{Comparison of RO-GQ($\lambda$), GQ($\lambda$), and LARS-TD on Triple-Link Inverted Pendulum Task showing minimum number of learning episodes.}
\label{table1}
\end{table}
\subsection{High-dimensional Under-actuated Systems}

The triple-link inverted pendulum \cite{si2001online} is a highly nonlinear under-actuated system with $8$-dimensional state space and discrete action space.  The state space consists of the angles and angular velocity of each arm as well as the position and velocity of the car. The discrete action space is $\{ 0,5{\rm{Newton}}, - 5{\rm{Newton}}\}$. The goal is to learn a policy that can balance the arms for $N_x$ steps within some minimum number of learning episodes. The allowed maximum number of episodes is $300$.
The pendulum initiates from zero equilibrium state and the first action is randomly chosen to push the pendulum away from initial state.
We test the performance of RO-GQ($\lambda$), GQ($\lambda$) and LARS-TD.
Two experiments are conducted with $N_x = 10,000$ and $100,000$, respectively. Fourier basis \cite{konidaris:fourier} with order $2$ is used, resulting in $6561$ basis functions.
Table 2 shows the results of this experiment, where RO-GQ($\lambda$) performs better than other approaches, especially in Experiment 2, which is a harder task. LARS-TD failed in this domain, which is mainly not due to LARS-TD itself but the quality of samples collected via random walk.

To sum up,  RO-GQ($\lambda$) tends to outperform GQ($\lambda$) in all aspects, and is able to outperform LARS-TD based policy iteration in high dimensional domains, as well as in selected smaller MDPs where LARS-TD diverges (e.g., the star MDP). It is worth noting that the computation cost of LARS-TD is $O(Ndp^3)$, where that for RO-TD is $O(Nd)$. If $p$ is linear or sublinear w.r.t $d$, RO-TD has a significant advantage over LARS-TD. However, compared with LARS-TD, RO-TD requires fine tuning the parameters of $\alpha_t, \rho_1, \rho_2$ and is usually  not as sample efficient as LARS-TD. We also find that tuning the sparsity parameter $\rho_2$ generates an interpolation between GQ($\lambda$) and  TD learning, where a large $\rho_2$ helps eliminate the correction term of TDC update and make the update direction more similar to the TD update.
\section{Conclusions}
This paper presents a novel unified framework for designing regularized off-policy convergent RL algorithms combining a convex-concave saddle-point problem formulation for RL with stochastic first-order methods. A detailed experimental analysis reveals that the proposed RO-TD algorithm is both off-policy convergent and is robust to noisy features.  There are many interesting future directions for this research.
One direction for future work is to extend the subgradient saddle-point solver to a more generalized mirror descent framework. Mirror descent is a generalization of subgradient descent with non-Euclidean distance \cite{ben2005non}, and has many advantages over gradient descent in high-dimensional spaces.
In \cite{sra2011optimization}, two algorithms to solve the bilinear saddle-point formulation are proposed based on mirror descent and the extragradient \cite{extragradient:1976}, such as the Mirror-Prox algorithm. \cite{sra2011optimization} also points out that the Mirror-Prox algorithm may be further optimized via randomization. %
To scale to larger MDPs, it is possible to design SMDP-based mirror-descent methods as well.

\section*{Acknowledgments}

This material is based upon work supported by the Air Force Office of Scientific Research (AFOSR) under
grant FA9550-10-1-0383, and the  National
Science Foundation under Grant Nos. NSF CCF-1025120,
IIS-0534999,  IIS-0803288, and IIS-1216467 Any opinions, findings, and conclusions or recommendations expressed
in this material are those of the authors and do not necessarily reflect the views of the AFOSR or the
NSF. We thank M. F. Duarte for helpful discussions.

\newpage
\bibliography{reference}
\bibliographystyle{plain}

\end{document}